\documentclass{article}
\usepackage{ijcai23}

\usepackage{times}
\usepackage{soul}
\usepackage{url}
\usepackage{scrextend}
\usepackage[hidelinks]{hyperref}
\usepackage[utf8]{inputenc}
\usepackage[small]{caption}
\usepackage{subcaption}
\usepackage{graphicx}
\usepackage{amsmath}

\usepackage{amsthm}

\usepackage{amsmath}
\usepackage{cleveref}
\crefformat{section}{\S#2#1#3} 
\crefformat{subsection}{\S#2#1#3}
\crefformat{subsubsection}{\S#2#1#3}

\usepackage{booktabs}
\usepackage{algorithm}
\usepackage{algorithmic}
\usepackage[switch]{lineno}

\newcount\Comments  
\usepackage{color}
\definecolor{darkgreen}{rgb}{0,0.5,0}
\definecolor{purple}{rgb}{1,0,1}
\newcommand{\kibitz}[2]{\ifnum\Comments=1\textcolor{#1}{#2}\fi}

\newcommand{\cbr}{Case-Based Reasoning}

\title{
\cbr~with Language Models for Classification of Logical Fallacies 
}

\author{
    \affiliations
    \emails
}

\author{
Zhivar Sourati$^{1,2}$
\and
Filip Ilievski$^{1,2}$
\and
Hông-Ân Sandlin$^{3}$
\And
Alain Mermoud$^3$
\affiliations
$^1$Information Sciences Institute, University of Southern California, Marina del Rey, CA, USA\\
$^2$Department of Computer Science, University of Southern California, Los Angeles, CA, USA\\
$^3$Cyber-Defence Campus, armasuisse Science and Technology, Switzerland\\
\emails
\{souratih,ilievski\}@isi.edu,
\{hongan.sandlin,alain.mermoud\}@ar.admin.ch
}

\begin{document}

\maketitle

\begin{abstract}
The ease and speed of spreading misinformation and propaganda on the Web motivate the need to develop trustworthy technology for detecting fallacies in natural language arguments. However, state-of-the-art language modeling methods exhibit a lack of robustness on tasks like logical fallacy classification that require complex reasoning. In this paper, we propose a Case-Based Reasoning method that classifies new cases of logical fallacy by language-modeling-driven retrieval and adaptation of historical cases. We design four complementary strategies to enrich input representation for our model, based on external information about goals, explanations, counterarguments, and argument structure. Our experiments in in-domain and out-of-domain settings indicate that Case-Based Reasoning improves the accuracy and generalizability of language models. Our ablation studies suggest that representations of similar cases have a strong impact on the model performance, that models perform well with fewer retrieved cases, and that the size of the case database has a negligible effect on the performance. Finally, we dive deeper into the relationship between the properties of the retrieved cases and the model performance. 
    
\end{abstract}

\section{Introduction}
\label{sec:introduction}

The ease and speed of spreading misinformation~\cite{wu2019misinformation,allcott2019trends} and propaganda \cite{da2019fine,barron2019proppy} on the Web motivate the need to develop trustworthy technology for understanding novel arguments~\cite{10.1162/coli_a_00364}. 
Inspired by centuries of philosophical theories~\cite{Aristotle1989-jz,Locke1997-co,Copi1954-COPITL-6,barker1965elements}, recent work has proposed the natural language processing (NLP) task of \textit{Logical Fallacy Detection}. Logical Fallacy Detection goes beyond prior work on binary detection of misinformation and fake news classification, and aims to classify an argument into one of the dozens of fallacy classes. For instance, the argument \textit{There is definitely a link between depression and drinking alcoholic drinks. 
I read about it from Wikipedia} belongs to the class \textit{Fallacy of Credibility}, as the validity of the argument is based on the credibility of the source rather than the argument itself.
Here, the focus is on informal fallacies that contain incorrect or irrelevant premises, as opposed to formal fallacies, which have an invalid structure~\cite{Aristotle1989-jz}.
The identification of informal fallacies is challenging for both humans and machines as it requires complex reasoning and also common knowledge about the concepts involved in the fallacy \cite{sep-fallacies}. To predict the correct fallacy type, the model has to know what Wikipedia is and how it is used in societal discourse, the potential relationship between depression and consuming alcoholic beverages, and also the causal link between the first and second parts of the argument.

The currently dominant NLP paradigm of language models (LMs) has been shown to struggle with reasoning over logical fallacies~\cite{logical_fallacy_main_paper} and similar tasks that require complex reasoning~\cite{da2019fine,barron2019proppy}. 
As LMs are black boxes, attempts to improve their performance often focus on adapting their input data. Prior work has pointed to the need to include context~\cite{vijayaraghavan_tweetspin_2022}, simplify the input structure~\cite{logical_fallacy_main_paper}, or perform special training that considers soft logic~\cite{https://doi.org/10.48550/arxiv.2002.05867}. However, these ideas have not been successful in classifying logical fallacies yet. 
Alternatively, methods that leverage reasoning by example, e.g., based on Case-Based Reasoning (CBR), have shown promise in terms of accuracy and explainability for other tasks like question answering~\cite{das2022knowledge}, but have not been applied to reason over logical fallacies to date. We conclude that integrating such explainable methods with generalizable LMs provides an unexplored opportunity to reason over logical fallacies.

In this paper, we pursue the question: 
\textit{
Does reasoning over examples improve the ability of language models to classify logical fallacies?
}
To answer this question, we develop a method based on the idea of CBR \cite{Aamodt1994}. We focus on the interpretive problem-solving variant of CBR, which aims to understand novel cases in terms of previous similar cases while not necessarily using the solutions from previous cases directly~\cite{LEAKE200112117}.
We adapt this idea to the task of classifying logical fallacies, by using LMs as backbones when retrieving and adapting prior similar cases. We measure the ability of our models in terms of accuracy and generalizability, and also probe their explainability. The main contributions of this paper are as follows:

\begin{enumerate}
    \item We design the first \textbf{\cbr~method} for logical fallacy classification to solve new cases based on past similar cases. The framework implements the theory of CBR with state-of-the-art (SOTA) techniques based on language modeling and self-attention. 
    \item We design four \textbf{enriched case representations}: \textit{Counterarguments}, \textit{Goals}, \textit{Explanations}, and \textit{Structure} of the argument to allow CBR to retrieve and exploit similar cases based on implicit information, like argument goals. To our knowledge, we are the first who investigate the effect of these case representations on CBR performance.
    \item We perform \textbf{extensive experiments} that investigate the impact of CBR against Transformer LM baselines
    on in-domain and out-of-domain settings. We perform ablations to provide insight into the sensitivity of our CBR method on its parameters and investigate the explanations extracted from the model.
\end{enumerate}
We make our code and data available to support future research on logical fallacy classification. \footnote{\url{https://github.com/zhpinkman/CBR}}

\section{Method}
\label{sec:method}

\begin{figure}[h]
    \centering
    \includegraphics[width=0.85\linewidth]{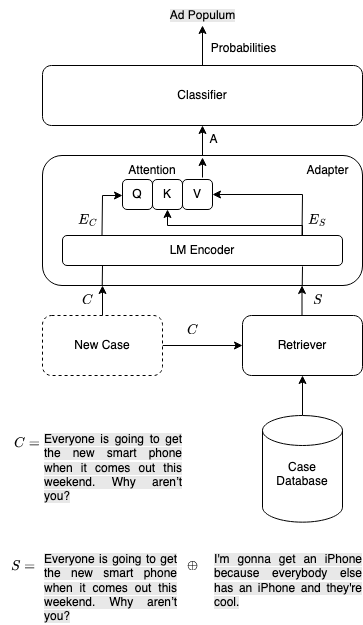}
    \caption{Three stages of the CBR pipeline. 
    Using the new case $C$, the retriever finds $k$ similar cases $\{ S_1, S_2, ..., S_k \}$ and creates $S = C\; \oplus <SEP> \oplus S_1 \oplus S_2 \oplus ... \oplus S_k $. The adapter processes both the new case and fetched similar cases and tries to adapt $S$ based on the new case $C$, and extracts more abstract information from the fusion of the two. Finally, the classifier receives the adapted information and returns the probabilities associated with the new class belonging to each fallacy type. In the example, $k = 1$.
    }
    \label{fig:cbr}
\end{figure}

CBR \cite{10.5555/538776} is a method that reasons over new cases based on similar past cases with a known label~\cite{Aamodt1994}. Our CBR formulation (Figure \ref{fig:cbr}) consists of three steps:
(1) given a new case, \textit{retrieve} similar cases from the case database, 
(2) \textit{adapt} the fetched similar cases based on the current one, and
(3) \textit{classify} the new case based on the adapted exemplars. 
In this work, we use LMs as key components in the retriever and the adapter. We opt for this choice because of their strong ability to encode and compute similarity for any natural language input.

\textbf{Retriever} 
finds $k$ similar cases $S_i\;\;(i \in \{1, ..., k\})$ to the new case $C$ from a case database. 
The retriever estimates the similarity between $C$ and $S_i$ by encoding each of them with the same LM encoder and computing the cosine similarity of the resulting encodings.
The retriever then picks the $k$ cases with top cosine similarities from the database. The new case is concatenated to its similar cases, i.e., $S = C \;\oplus <SEP> \oplus S_1 \oplus S_2 \oplus ... \oplus S_k $ and is passed as input to the CBR adapter.


\textbf{Adapter} aims to prioritize the most relevant information from $S$ for reasoning over the new case $C$.
Based on the second step of the pipeline by \cite{Aamodt1994}, after fetching similar cases, it might be the case that only certain retrieved cases would be useful, and therefore, they should be weighted according to their utility for approaching the new case. The fusion of the current case with its previously seen similar problems would give the model the chance to come up with a better representation of the current problem, as well as better abstractions and generalizations for further uses.

The adapter consists of two parts: an \textit{encoder} and an \textit{attention component}. 
The encoder is an LM that takes as an input $C$ and $S$ separately, then outputs their respective embedding representations $E_C$ and $E_S$. We use the hidden states of the last layer of the LM as the input embedding. 
A multi-headed attention component \cite{https://doi.org/10.48550/arxiv.1706.03762} with $H$ heads selects the most useful information from the similar cases embeddings $E_S$ given the embedding of the new case $E_C$. As commonly done in Transformer architectures, the Adapter generates Value and Key vectors from $E_S$ and Query vectors from $E_C$. The dot product of the Query and Key vectors, fed through a softmax layer, results in an Attention vector, which indicates the importance of each token in $S$ when generating the adapted vector $A$. An adapted vector with adjusted attention on its elements is produced by the weighted sum of the Value vectors based on Attention weights. The output of the attention component is $A$, the adjusted embedding of $E_S$. 

\textbf{Classifier} predicts the final class based on the adapter output $A$. The classifier is designed as a fully connected neural layer with a depth $d$ and an activation function. The objective function of the classifier is the cross-entropy loss. The cross-entropy loss is computed over the probabilities that are extracted from $C$ logits that correspond to each of the $C$ classes. Also, during training, the retriever's component weights are frozen, while the adapter and the classifier are trained in an end-to-end fashion. 

Overall, our CBR architecture resembles a standard `vanilla' LM with a classification head but brings the additional benefit of having access to prior relevant labeled cases weighed based on the attention mechanism.\footnote{Our experiments using the framework without the attention mechanism consistently showed sub-optimal performance.}
We hypothesize that the CBR models bring two benefits over vanilla LMs:
(1) the integration of similar labeled cases helps the model analyze the new fallacious argument better and classify it more accurately, and (2) provides explicit insights into the reasoning of the model by yielding similar cases to the current one~\cite{https://doi.org/10.1111/cogs.12086}.

\section{Case Representation}
\label{sec:case-representation}

Merely retrieving labeled cases may not be sufficient for reasoning on new cases, as it is unclear what dimensions of similarity their relevance is based on. 
For instance, two cases may be similar in terms of their explanation, structure, or the goal behind the cases. As these dimensions are implicit and not apparent from the plain text, we make them explicit by enriching the original text of the case with such information. We consider four representations in which the case formulation is enriched with its \textit{counterargument}, \textit{goal}, \textit{explanation}, and \textit{structure}.
As a baseline, we also include the original \textit{text} without any enrichments.
Table \ref{tab:case-representations} illustrates examples of these representations for the sample case \textit{There was a thunderstorm with rain therefore I did not finish my homework}.

Each of the enrichment strategies $r$ modifies the case representation by concatenating it with additional information, $r(case)$. We introduce a case representation function $R(case, r)$ that concatenates $case$ with additional information $r(case)$ resulting in $case\oplus r(case)$. These representations modify both the new case $C$ to $R(C,r)$ and cases from the database $S_i$ to $R(S_i,r)$, and change the cosine similarity to be computed between enriched cases instead of plain text. We next describe the design of the enrichment strategies.

\begin{table*}[h]
\centering
\small
\begin{tabular}{p{0.12\textwidth} | p{0.82\textwidth}}
\toprule

\textbf{Representation}  &  \textbf{Transformed Text} \\
\midrule
\textit{Goals} & It's possible that the goal is to explain why the speaker did not finish their homework. The speaker may be trying to convince the listener that they did not finish their homework because of the thunderstorm. \\
\midrule
\textit{Counterarg.} & There are many factors that contribute to a person's ability to complete their homework, and it's not fair to suggest that the thunderstorm was the only factor. It's possible that the person did not finish their homework because they were distracted by the thunderstorm or because they were tired. \\
\midrule
\textit{Explanations} & It presents a causal relationship between two events that might not be actually related. \\
\midrule
\textit{Structure} & There was an X with Y therefore I did not do Z. \\
\bottomrule
\end{tabular}
\caption{
One example of different representations for the case \textit{There was a thunderstorm with rain therefore I did not finish my homework}.
}
\label{tab:case-representations}
\end{table*}

\textbf{Counterarguments.} 
Counterarguments are common in persuasive writing, where they explain why one's position is stronger than the counterargument and serve as a preemptive action to anticipate and remove any doubts about arguments \cite{harvey_1999}.
We hypothesize that counterarguments are often implicit in the arguments, and would therefore be useful to be provided directly to the model.
For instance, in the argument presented in Table \ref{tab:case-representations}, although the plain text claims that the reason for not finishing the homework is the heavy rain, the counterargument points out \textit{other reasons for not finishing the homework such as the person being too tired}. 

\textbf{Goals.}
Studies of argumentation often focus on the interplay between the goals that the writer is pursuing and their argumentations \cite{tracy2013understanding}. Thus, when classifying logical fallacies, we expect that it is beneficial to take into account the goals of the arguments. The goal may be entirely missing in the argument's text, or the argument may implicitly hint at the goal. An example of the latter is shown in Table \ref{tab:case-representations}, where the phrase \textit{therefore I did not finish my homework} alludes to the implicit goal of the writer to \textit{justify not finishing their homework}. As shown in this example, we include an explicit goal statement to fill this gap.

\textbf{Explanations.} 
By using explanations about logically fallacious arguments, we aim to augment the arguments with a broader notion of information that might be useful for classifying logical fallacies but is not already included in the original argument, such as reasoning steps getting from premises to conclusions of an argument \cite{barker1965elements}. 

As we do not impose any restrictions on the explanations, their content may overlap with the previous two representations. Alternatively, explanations may provide different complementary information. Such is the example in Table~\ref{tab:case-representations} that discusses the \textit{causal relationship between two events that are not actually related}. Thus, the explanation acts as a general gap-filling mechanism that can provide any relevant information that is missing in the original argument.

\textbf{Structure.}
Tasks like logical fallacy classification involve higher-order relation comprehension that is often based on the structure rather than the content of the argument. In that sense, the semantics of specific entities and concepts in the argument may be misleading to the model. Similarly to \cite{logical_fallacy_main_paper}, we hypothesize that focusing on the logical structure of an argument rather than its content is beneficial for the model's performance~\cite{gabbay2004handbook}. An example of a structural simplification of an argument is presented in Table \ref{tab:case-representations}. While this simplification may help the model grasp the case structure more directly, the structure formulation may not detect the implicit causal links between the thunderstorm (X) and the homework (Z).

We extract the enrichment information for a case using a combination of few-shot and zero-shot prompting with two SOTA models: ChatGPT \cite{openai_2023} and Codex \cite{https://doi.org/10.48550/arxiv.2107.03374}. Given a representation strategy $r$, we prompt ChatGPT to get the representations for a case for five different examples using one template per representation. For instance, we use the template \textit{Express the goal of the argument $\{case\}$} to retrieve the \textit{goals} of the argument $case$. The full list of the templates is provided in Appendix. 
The five obtained examples per representation are used as demonstrations to prompt Codex in a few-shot manner. For a representation strategy $r$, we use the same demonstrations together with each new case $C$ from our task as input to the Codex model, which yields enrichment information $r(case)$ per case. In this manner, we combine the strong zero-shot ability of the closed-source ChatGPT model with the few-shot generation strength of the Codex model.

\section{Experimental Setup}
\label{sec:experimental-setup}

In this section, we describe the evaluation data and metrics, the baselines we compare to, and the implementation details.

\textbf{Evaluation Dataset.}
We use two logical fallacy datasets from \cite{logical_fallacy_main_paper}, called LOGIC and LOGIC Climate. The LOGIC dataset includes thirteen logical fallacy types about common topics, namely:
\textit{Ad Hominem, Ad Populum, Appeal to Emotion, Circular Reasoning, Equivocation, Fallacy of Credibility, Fallacy of Extension, Fallacy of Logic, Fallacy of Relevance, False Causality, False Dilemma, Faulty Generalization,} and \textit{Intentional} (examples and statistics provided in Appendix).
LOGIC Climate dataset consists of more challenging examples for the same logical fallacy types on the climate change topic. We use LOGIC for in-domain evaluation and LOGIC Climate for out-of-domain evaluation. As these datasets are severely imbalanced, we augment them using two techniques, i.e., back-translation, and substitution of entities in the arguments with their synonymous terms. This augmentation makes the dataset have 281 arguments for each fallacy type (find a more in-depth discussion of augmentation in Appendix).
Note that we do not fine-tune our model on the LOGIC Climate dataset in any of our experiments to evaluate the generalizability of our framework. We model the classification task as a multi-class classification problem and use the customary metrics of weighted precision, recall, and F1-score.

\textbf{Baselines.}
We consider three different LMs: BERT \cite{bert}, RoBERTa \cite{roberta}, and ELECTRA \cite{electra}. We apply our CBR method (\cref{sec:method}) on each of these models. As baselines, we use vanilla LMs without a CBR extension. We also compare against Codex in a few-shot setting, with the prompt including all the possible classes as well as one example for each class resulting in thirteen labeled examples in the prompt (discussed more in detail in Appendix).
Finally, we include the results of a frequency-based predictor that predicts fallacy classes based on the distribution of fallacy types in the training set.

\textbf{Implementation details.} We use SimCSE \cite{https://doi.org/10.48550/arxiv.2104.08821}, a transformer-based retriever that is optimized for capturing overall sentence similarity, to compute the similarity between cases (\cref{sec:method}) and also use $H = 8$ heads for the multi-headed attention component.  The depth of our classifier is $d=2$. It uses $gelu$ \cite{https://doi.org/10.48550/arxiv.1606.08415} as an activation function.
We analyze the performance of our model using $k \in \{1,2,3,4,5\}$. To test the generalization of our model with sparser case databases, we experiment with various ratios of the case database within $\{0.1, 0.4, 0.7, 1.0\}$. 

\section{Results}
\label{sec:results}

In this section, we measure the effectiveness of CBR per model and case representation. We further provide ablations that measure the sensitivity of the model to the size of the case database and the number of cases. Finally, we present a qualitative analysis of the explainability of CBR and a thorough discussion about how retrieved cases help to classify new ones.

\begin{table}[t!]
\centering
\small
\resizebox{\columnwidth}{!}{
\begin{tabular}{@{}llrrrrrr@{}}
\toprule
& &
\multicolumn{3}{c}{ \textbf{LOGIC}} & \multicolumn{3}{c}{ \textbf{LOGIC Climate}} \\
\cmidrule(lr){3-5} \cmidrule(lr){6-8}
\textbf{Model} & \textbf{Type} & P & R & F1  & P & R & F1\\
\midrule
Freq-based & baseline & 0.094 & 0.094 & 0.093  & 0.120 & 0.079 & 0.080 \\
Codex & few-shot &  0.594 & 0.422 & 0.386  & 0.198 & 0.093 & 0.077 \\
\midrule
ELECTRA & baseline & 0.614 & 0.602 & 0.599  & 0.276 & 0.229 & 0.217  \\
& CBR & \underline{\textbf{0.663}} & \underline{\textbf{0.664}} & \underline{\textbf{0.657}} & \textbf{0.355} & \underline{\textbf{0.254}} & \underline{\textbf{0.270}} \\
\midrule
RoBERTa & baseline & 0.577 & 0.561 & 0.560 & 0.237 & 0.211 &	0.200 \\
& CBR & \textbf{0.631} & \textbf{0.619} & \textbf{0.619} & \underline{\textbf{0.379}} & \textbf{0.248} & \textbf{0.245} \\
\midrule
BERT & baseline & 0.585 &	0.598 & 0.586 & 0.166 & 0.130 & 0.120 \\
& CBR & \textbf{0.613}	& \textbf{0.616}	& \textbf{0.611}	& \textbf{0.359}	& \textbf{0.204}	& \textbf{0.200} \\
\bottomrule
\end{tabular}
}
\caption{
Comparison of the best results of the CBR framework with vanilla LMs and two external baselines on two benchmarks focusing on both in-domain (LOGIC) and out-of-domain (LOGIC Climate) settings. The best results per model are \textbf{boldfaced} and the overall best results are \underline{underlined}.
}
\label{tab:main_results}
\end{table}

\textbf{The Impact of CBR.}
Table \ref{tab:main_results} shows the performance of the CBR framework and relevant baselines. For each model, we present the results using the best case representation per model and using $k = 1$ while exploiting 10\% of the case database that we found to yield the best results among all possible combinations. 

Overall, the CBR method brings a consistent and noticeable quantitative improvement in the classification of logical fallacies by LMs. For each of the three LMs, CBR outperforms the vanilla baselines by 2.5 - 6 absolute F1 points on the in-domain dataset and up to 8 points on the out-of-domain dataset. Furthermore, CBR outperforms Codex, which is utilized in a few-shot setting, despite it being a much larger model. Across the different LMs, ELECTRA is achieving the best score and benefits the most from the CBR framework on the in-domain benchmark, which we attribute to its efficiency of pre-training \cite{electra}. The same pattern of the superiority of CBR over vanilla LMs can be observed for the other two models with different pre-training procedures and varying numbers of internal parameters. The CBR method notably and consistently improves the performance of the LMs on the out-of-domain (LOGIC Climate) benchmark as well, with ELECTRA performing the best and BERT benefiting the most from CBR. We provide detailed per-class results of our method in the Appendix, demonstrating its ability to improve the accuracy of the baseline for each of the thirteen fallacy classes. Especially we note that our CBR method is able to increase the performance in the classes with the least examples, such as Equivocation (0 $\rightarrow$ 0.35) and Fallacy of Extension (0.48 $\rightarrow$ 0.75).

We conclude that CBR is a general framework that can be applied to any LM and can generalize well to unseen data and to various fallacy classes. The generalization of CBR is in line with prior work that suggests its strong performance on tasks with data sparsity \cite{https://doi.org/10.48550/arxiv.2006.14198}.

\begin{table}[h]
\centering
\small
\resizebox{\columnwidth}{!}{%
\begin{tabular}{@{}llrrrrrr@{}}
\toprule
& &
\multicolumn{3}{c}{ \textbf{LOGIC}} & \multicolumn{3}{c}{ \textbf{LOGIC Climate}} \\
\cmidrule(lr){3-5} \cmidrule(lr){6-8}
\textbf{Model} & \textbf{Representation} & P & R & F1  & P & R & F1\\
\midrule
ELECTRA  
& \textit{Text}         &     0.655 &  0.634 &  0.635 &             0.317 &          0.242 &      0.242 \\
& \textit{Counterarg.}      &     \underline{\textbf{0.663}} &  \underline{\textbf{0.664}} &  \underline{\textbf{0.657}} &             0.355 &          \underline{\textbf{0.254}} &      \underline{\textbf{0.270}} \\
& \textit{Goals}        &     0.646 &  0.622 &  0.621 &             \textbf{0.376} &          0.217 &      0.222 \\
& \textit{Structure}    &     0.634 &  0.625 &  0.618 &             0.375 &          0.254 &      0.269 \\
& \textit{Explanations} &     0.605 &  0.580 &  0.578 &             0.314 &          0.242 &      0.237 \\
\midrule
RoBERTa 
& Text         &     \textbf{0.633} &  0.613 &  0.619 &             0.343 &          0.236 &      0.251 \\
& \textit{Counterarg.}      &     0.624 &  0.613 &  0.615 &             0.367 &          0.198 &      0.216 \\
& \textit{Goals}        &     0.632 &  0.613 &  0.619 &             0.351 &          0.242 &      \textbf{0.263} \\
& \textit{Structure}    &     0.631 &  \textbf{0.619} &  \textbf{0.619} &             \underline{\textbf{0.379}} &          \textbf{0.248} &      0.245 \\
& \textit{Explanations} &     0.575 &  0.558 &  0.559 &             0.359 &          0.192 &      0.181 \\
\midrule
BERT 
& \textit{Text}         &     0.595 &  0.604 &  0.596 &             0.311 &          0.192 &      0.204 \\
& \textit{Counterarg.}      &     0.607 &  0.613 &  0.603 &             0.342 &          \textbf{0.217} &      \textbf{0.228} \\
& \textit{Goals}        &     0.598 &  0.607 &  0.596 &             0.310 &          0.204 &      0.203 \\
& \textit{Structure}    &     \textbf{0.613} &  \textbf{0.616} &  \textbf{0.611} &             \textbf{0.359} &          0.204 &      0.200 \\
& \textit{Explanations} &     0.540 &  0.531 &  0.532 &             0.274 &          0.217 &      0.190 \\
\bottomrule
\end{tabular}}
\caption{
Performance of the CBR framework using different case representations. The best results per model are \textbf{boldfaced} and the overall best results are \underline{underlined}. 
}
\label{tab:results-representations}
\end{table}

\textbf{Effect of Different Representations.}
The results in Table \ref{tab:results-representations} confirm our expectation that the case representation plays an important role in the effectiveness of the CBR framework. Depending on the LM used, the performance difference among different case representations ranges from 6 to 8\% F1-scores for the in-domain setting and 4 to 8\% F1-scores for the out-of-domain setting.
In general, we observe a boost in performance when enhancing the original representation (\textit{text}). \textit{Counterargument} information yields the highest boost, though the impact of the representations varies across models. Using ELECTRA, the enrichment with \textit{counterarguments} helps the most, outperforming the model based on the original \textit{text} and the other enrichment strategies. With RoBERTa, \textit{goals} and \textit{structure} of the arguments perform on par with \textit{text}, while with BERT, including information about \textit{counterarguments} and argument \textit{structure} outperforms the \textit{text} representation. 
As the LMs have been trained with different data and may optimize for different notions of similarity, it is intuitive that the impact of the case representations varies across models. This finding is in line with theoretical work, which discusses that knowledge transfer is strictly guided by the similarity function of the reasoning model~\cite{holyoak1996mental}. Meanwhile, using a generic enrichment with \textit{explanations} performs consistently poorly and harms the model performance, which suggests that the CBR models benefit from more precise case representations.

\begin{figure}
     \centering
     \begin{subfigure}[b]{0.5\columnwidth}
         \centering
         \includegraphics[width=\linewidth]{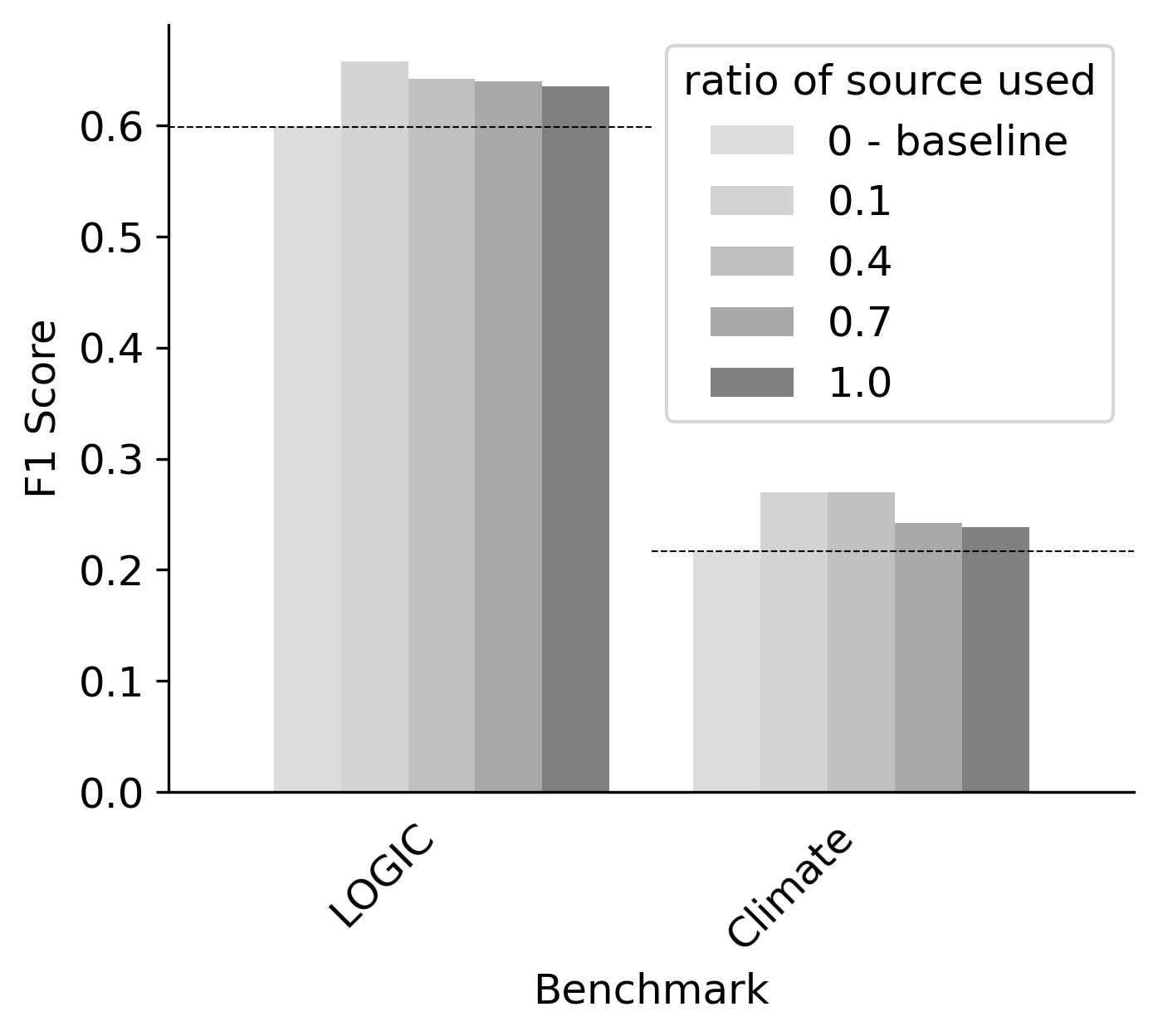}
         \caption*{}
         \label{fig:ratio-of-source-used}
     \end{subfigure}
     \hfill
     \begin{subfigure}[b]{0.48\columnwidth}
         \centering
         \includegraphics[width=1.0\linewidth]{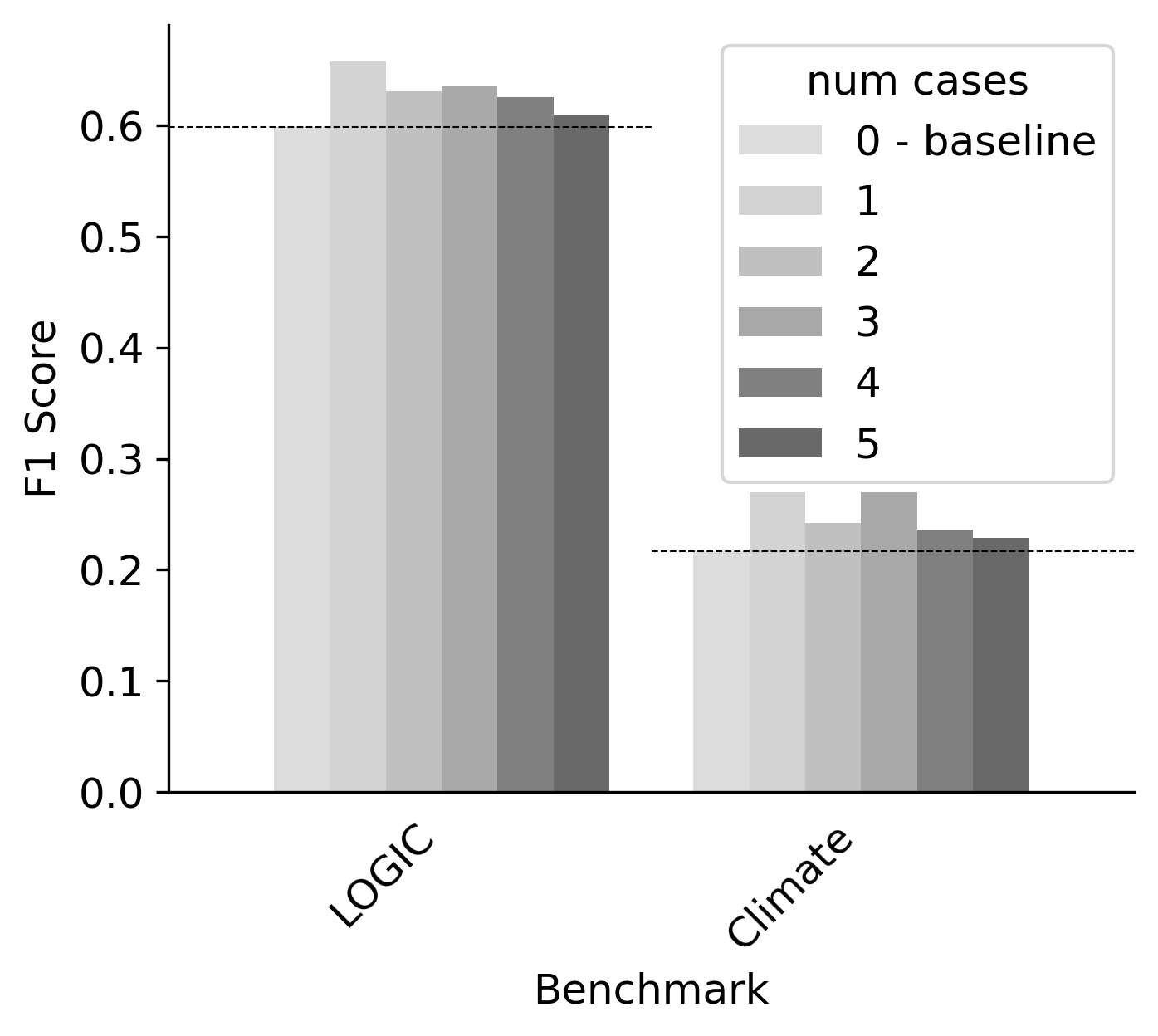}
         \caption*{}
         \label{fig:num-cases}
     \end{subfigure}
    \vspace*{-7mm}
    \caption{
    Performance of the CBR framework using different ratios of the case database (left) and different numbers of cases (right). The baseline is outlined as the dotted line. 
    }
    \label{fig:agg}
\end{figure}

\textbf{Effect of the Case Database Size.}
Next, we investigate the sensitivity of the best-performing CBR model based on ELECTRA to the size of the case database. Figure \ref{fig:agg} (left) depicts the performance of this model using different ratios of the case database. The figure shows that the CBR framework consistently outperforms the vanilla LM baseline (with 0\% of cases) on in- and out-of-domain settings. This trend stands regardless of the size of the case database, which indicates the low sensitivity of the CBR model to the case database size. However, we note that using 10\% of the case database yields the best performance, which indicates that a limited case database offers a better potential of abstraction to CBR. Moreover, comparing the performance of the model using different ratios of the case database, we observe a continuous decrease in the performance using higher percentages of the case database. Having access to too much data makes the model dependent and sensitive to the unnecessary and insignificant details of similar cases retrieved. These observations point us to the data efficiency properties of the CBR framework \cite{https://doi.org/10.48550/arxiv.2006.14198}.

\textbf{Effect of Different Number of Cases.} 
The performance of the best CBR model that uses ELECTRA with different numbers of cases is illustrated in Figure \ref{fig:agg} (right). In both in-domain and out-of-domain settings, we observe a consistent pattern of performance decrease when more cases are taken into account in the reasoning process. The CBR framework reaches its peak performance using only one similar case while once more outperforming the vanilla LM (with 0 cases) in all the settings. 
This indicates that the models get easily overwhelmed with past information when considering a new case. While intuitively, one would expect that a larger number of cases should help the model analyze a new case better, the reasoner should have the capacity to process all of these past cases. Otherwise, as observed in this experiment, including more cases can have an adverse effect on the reasoner by distracting it rather than helping it.

\begin{table*}[h]
\centering
\small
\begin{tabular}{p{0.18\textwidth} | p{0.3\textwidth} | p{0.35\textwidth}| p{0.075\textwidth} }
\toprule
 \textbf{Input Sentence} & \textbf{Enriched Representation for Correct Prediction (representation)} & \textbf{Enriched Representation for Wrong Prediction (representation) (predicted class)} & \textbf{Class}\\
\midrule
 People who don't support the proposed minimum wage increase hate the poor. & There are often multiple perspectives on an issue. It's possible to have a nuanced or balanced view that doesn't align with any side completely. \textit{\textbf{(Counterarg.)}} & That candidate wants to raise the minimum wage, but they aren't even smart enough to run a business. \textit{\textbf{(Text)}} \textit{\textbf{(Ad Hominem)}} & \textit{Fallacy of Extension}\\
\midrule
 The house is white; therefore it must be big. & X is y; therefore, it is z. \textit{\textbf{(Structure)}} & The sentence "People who drive big cars hate the environment" presents a generalization about a group of people without sufficient evidence and it relies on oversimplification.\textit{\textbf{(Explanations)}} \textit{\textbf{(Faulty Generalization)}} & \textit{Fallacy of Logic}\\ 
\midrule
Student: You didn't teach us this; we never learned this. Teacher: So, you're either lazy or unwilling to learn is that right? & It's possible that the argument "It’s possible to pass the class without attending. so, you will pass even if you don’t attend" is trying to convince the listener that they will pass the class even if they don't attend. The speaker may be trying to persuade the listener to skip class. \textit{\textbf{(Goals)}} & The sentence "Teacher: You are receiving a zero because you didn't do your homework. Students: Are you serious? You gave me a zero because you hate me?" attacks the person making the argument rather than the argument itself. \textit{\textbf{(Explanations)}} \textit{\textbf{(Fallacy of Extension)}} & \textit{False Dilemma} \\
\midrule
One day, Megan wore a Donald Duck shirt, and she got an A on her test. Now she wears that shirt every day to class. & There are many factors that contribute to a student's grade, and it's not fair to suggest that the student's past grades are the only factor. It's possible that the student failed the test because they didn't study, or because they were sick. \textit{\textbf{(Counterarg.)}} & The sentence "Eating five candy bars and drinking two sodas before a test helps me get better grades. I did that and got an A on my last test in history" presents a causal relationship between two events without sufficient evidence to support the claim. \textit{\textbf{(Explanations)}} \textit{\textbf{(Fallacy of Relevance)}} & \textit{False Causality} \\
\bottomrule
\end{tabular}
\caption{
Four examples from different classes in which the CBR model predicts the correct class. For each example, we show a representation that leads to a correct prediction and a representation that still leads to predicting the wrong class. We also show the corresponding wrong class predicted by the second variation of the model.
} 
\label{tab:cbr-examples}
\end{table*}

\textbf{Case Study on Explainability.} 
A key promise of the CBR framework is its native explainability by cases since its retrieval of similar cases and reasoning over them are integrated into the CBR process. We perform a qualitative analysis of the cases retrieved by CBR to develop a better intuition about its reasoning process. Table \ref{tab:cbr-examples} illustrates four example cases that the vanilla LM classifies incorrectly. For each case, we show two CBR representations: one leading to a correct prediction and one leading to an incorrect one.

The first example shows the scenario where the original \textit{text} of a retrieved case does not suffice for the model to reason correctly, despite its topical surface similarity to the input case. In other words, the high surface similarity of similar cases is confusing the model and forcing it to incorrectly predict the same class that is associated with the retrieved similar case. However, we see that enriching the case with its \textit{counterargument} helps the CBR model, even though the counterargument is phrased in an abstract manner and is not similar to the new case on the surface. 
We observe a similar situation with the \textit{explanations} enrichment in the third example, having high surface similarity between the retrieved case and the new one, where analyzing the argument \textit{goals} instead helps the model. 
In the second example, the \textit{structure} of the argument and the logical depiction of the past cases help the most, while in the fourth example, the \textit{counterarguments} assist the reasoning of CBR. 
From the second and the fourth example, we observe that enriching arguments with cases that are semantically far from the new case is confusing for the CBR model, even if their reasoning would be helpful.

\begin{table}[h]
\centering
\small
\begin{tabular}{p{1.9cm}p{1.1cm}p{1.2cm}p{1.1cm}p{1.2cm}}
\toprule
&
\multicolumn{2}{c}{ \textbf{LOGIC}} & \multicolumn{2}{c}{ \textbf{LOGIC Climate}} \\
\cmidrule(lr){2-3} \cmidrule(lr){4-5}
\textbf{Representation} &  \textbf{ground truth overlap} &  \textbf{predictions overlap} &  \textbf{ground truth overlap} &  \textbf{predictions  overlap} \\
\midrule
\textit{Text} & 0.184 &   0.232 &    \textbf{0.136} &   0.173 \\
\textit{Counterarg.} & 0.208 & 0.220 &   0.062 &    0.068 \\
\textit{Goals} & 0.178 & 0.196 &    0.130 &   0.124 \\
\textit{Structure} & 0.238 &   0.250 &   0.105 &   0.242 \\
\textit{Explanations} &\textbf{0.277} &   \textbf{0.447} &  0.086 &     \textbf{0.478} \\
\bottomrule
\end{tabular}
\caption{
Overlap of retrieved cases' labels with true labels and predictions of the best CBR model (ELECTRA). We highlight the highest overlaps in \textbf{bold}.
}
\label{tab:overlaps}
\end{table}

In summary, presented examples show that the retrieved cases help the model indirectly by providing CBR with high-level information (first example), symbolic abstractions (second example), extensive analysis of the writer's goal (third example), and alternative possibilities (fourth example). This brings up a natural question: does CBR performance correlate to class overlap between the current case and retrieved similar cases? In other words, can we label a new case solely based on its k-nearest neighbors' labels?
To answer this question, we compute the overlap of retrieved cases' labels with both the true and the predicted label for different case representations (Table \ref{tab:overlaps}).
We observe a low overlap of a maximum of 27.7\% between the retrieved cases' labels and the true labels, which is only slightly better than a frequency-based prediction. 
Also, centering on the direct effect of retrieved cases on the CBR predictions, the model with the highest class overlap between the retrieved cases and the predicted classes also has the lowest performance (\textit{explanations}). 
Meanwhile, the best CBR variants (e.g., \textit{counterarguments}) do not directly reuse the labels of the retrieved cases. We conclude that while retrieving similar cases provides the CBR models with useful information, this additional evidence influences the model reasoning indirectly and may have adverse effects otherwise. Although CBR, in its simplest form, can act as a k-nearest neighbors algorithm, our results suggest that the neighbors' labels cannot be used blindly, and further reasoning step over the retrieved cases is necessary. We believe that these findings open exciting future research directions that investigate the relationship between case similarity and CBR performance.

\section{Related Work}

In this section, we present prior research on logical fallacy classification, CBR, and methods that prompt very large LMs. 

\textbf{Logical Fallacy.}
Prior computational work on logical fallacies has mostly focused on formal fallacies using rule-based systems and theoretical frameworks \cite{Ireneous_Nakpih_2020}. Nevertheless, recent work has switched attention to informal logical fallacies and natural language input. \citeauthor{logical_fallacy_main_paper} propose the task of logical fallacy classification, considering thirteen informal fallacy types and two benchmarks. 
The authors gather a rich set of arguments containing various logical fallacies from online resources and evaluate the capabilities of large LMs in classifying logical fallacies both in in-domain and out-of-domain settings. Similarly, \citeauthor{goffredo2022fallacious} present a dataset of political debates from U.S. Presidential Campaigns and use it to evaluate Transformer LMs. Processing different parts of arguments, such as dialogue's context, they create separate expert models for each part of arguments and train all the models together, from which they report the importance of discussion context in argument understanding. Although LMs have been used to classify logical fallacies, both independently and in an ensemble setting, to our knowledge, no prior work has tried to improve LMs' capabilities to reason over previous cases of logical fallacies encountering a new case nor experimented with enriching the argument representation. We fill this gap by employing CBR with LMs to reason over similar past cases to classify logical fallacies in new cases.

\textbf{\cbr.} 
\cbr~\cite{10.5555/538776} has been a cornerstone of interpretable models in many areas. For instance, researchers have applied CBR over past experiences in mechanical engineering \cite{qin_regli_2003} and medical applications \cite{OYELADE2020100395}. \cbr~has been also used in education, particularly to teach students to recognize fallacies ~\cite{spensberger2022effects}. Exploiting its interpretable properties, \citeauthor{8776909} use \cbr~as a transparent model for Word Sense Disambiguation, \citeauthor{bruninghaus2006progress} use \cbr~ for predicting legal cases an interpretable pipeline, while \citeauthor{https://doi.org/10.48550/arxiv.2009.06349} use \cbr~ to enhance the transparency of classifications made on written digits \cite{726791}. Inspired by its advantages, we couple \cbr~ with LMs, leading to enhanced accuracy and explainability of classifying logical fallacies. To our knowledge, this is the first work that combines CBR with LMs for complex tasks like logical fallacy classification. Nevertheless, there are frameworks that are close to CBR that also have a notion of memory, but cannot serve as replacements, given their restrictions. Analogical reasoning \cite{GENTNER2012130} methods typically focus on proportions between words or short text sequences and cannot generalize well to unstructured text. K-nearest neighbor methods are a simplified version of CBR that, given our observations, can not perform as well as CBR. Our framework can also be seen broadly as a memory-based model \cite{weston2015memory}, however, our proposed formulation that combines CBR and LMs has not been explored before for tasks like logical fallacy classification. 

\textbf{Prompting LMs.} 
The behavior of LMs is dependent on the quality of their inputs. Aiming to create more comprehensive inputs for LMs and assist them in complex reasoning tasks, researchers have attempted to transfer knowledge from very large LMs to smaller ones. \citeauthor{selftalk} show that LMs can discover useful contextual information about the question they answer, from another LM. \citeauthor{pinto} propose an LM pipeline that learns to faithfully reason over prompt-based extracted rationales. 
\citeauthor{chothr} explore how generating a series of intermediate reasoning steps using prompting can equip LMs with complex reasoning skills. 
Inspired by the ability of large LMs to provide relevant information for novel inputs, as well as prior work that performs knowledge distillation from large to smaller LMs \cite{https://doi.org/10.48550/arxiv.2110.07178}, we use prompting to enrich the arguments containing logical fallacies. According to \cite{barker1965elements}, logical fallacies are created by transition gaps from premises to conclusions, and we try to enrich the arguments using prompting to cover the gaps. Our method resembles retrieval-augmentation methods \cite{lewis2021retrievalaugmented}, yet, our enrichment strategies are novel and have not been explored on such complex tasks.

\section{Conclusions and Future Work}

In this paper, we presented a novel method that uses \cbr~ with LMs to classify logical fallacies. The CBR method reasons over new cases by utilizing past experiences. To do so, the method retrieves the most relevant past cases, adapts them to meet the needs of a new case, and finally classifies the new case using the adjusted information from past cases. We devised four auxiliary case representations that enrich the cases with implicit information about their counterarguments, goals, structure, and explanations.
Our results showed that CBR can classify logical fallacies and can leverage past experiences to fill the gaps in LMs. CBR outperformed the LM baselines in all settings and across all thirteen logical fallacy classes.
CBR was able to generalize well and transfer its knowledge to out-of-domain setting. The representation of its cases played a key role: enriching cases with counterarguments helped the most, while adding generic explanations harmed the model's performance. Furthermore, CBR models performed best when a small number of cases are provided, but showed low sensitivity to the size of the case database. Finally, our qualitative analysis demonstrated the value of CBR as an interpretable framework that benefits from past similar cases indirectly.

Since our experiments showed that similar cases assist CBR indirectly, future research should further qualify the relationship between the information provided by the retrieved cases and the performance of the model.
Moreover, future work should focus on evaluating CBR on other natural language tasks that require abstraction, such as propaganda detection and dialogue modeling. For instance, given a task-oriented dialogue about cooking a new meal, the model may benefit from procedures for cooking similar meals. The application of CBR on such tasks might also inspire additional case enrichment strategies, e.g., that describe the causal relation between text chunks, and point to additional knowledge gaps that CBR needs to fill.


\section*{Acknowledgements}
Zhivar Sourati has been supported by armasuisse Science and Technology, Switzerland under contract No. 8003532866, and NSF under Contract No. IIS-2153546, while Filip Ilievski is sponsored by the DARPA MCS program under Contract No. N660011924033 with the US Office Of Naval Research.

\bibliographystyle{named}
\bibliography{ijcai23}


\appendix

\section{Dataset}
\label{appendix:dataset}

We start with the dataset from \cite{logical_fallacy_main_paper} that is further revised and cleaned by the authors.\footnote{\url{https://github.com/tmakesense/logical-fallacy/tree/main/dataset-fixed}} It consists of thirteen fallacy types presented in Table \ref{tab:fallacy-examples}.

\begin{table*}[!t]
\centering
\small
\begin{tabular}{p{0.16\textwidth}| p{0.35\textwidth}| p{0.45\textwidth}}
\toprule

\textbf{Fallacy Type} & \textbf{Definition} & \textbf{Example}\\
\midrule
\textit{Ad Hominem} & Instead of addressing someone's argument or position, you irrelevantly attack the person or some aspect of the person who is making the argument. & I don't know how Professor Resnick can be such a hard grader.  He's always late for class. \\
\midrule

\textit{Ad Populum} & A fallacious argument which is based on affirming that something is real or better because the majority thinks so. & An increasing number of people are keeping ferrets as pets, so they must make wonderful companion animals. \\
\midrule
\textit{Appeal to Emotion} & Manipulation of the recipient's emotions in order to win an argument. & If you don't buy the warranty for your chromebook, you could find yourself without one for the rest of the schoolyear and having to pay thousands of dollars in expensive repairs. \\
\midrule
\textit{Fallacy of Extension} & Attacking an exaggerated or caricatured version of your opponent's position. & Scientist: Evolution explains how animals developed, adapted and diversified over millions of years. Opponent: If we evolved from monkeys, why are there still monkeys? And why don't we have three arms? Wouldn't that give me a competitive advantage? \\
\midrule
\textit{Fallacy of Relevance} & These fallacies appeal to evidence or examples that are not relevant to the argument at hand.  & Grading this exam on a curve would be the most fair thing to do. After all, classes go more smoothly when the students and the professor are getting along well. \\ 
\midrule
\textit{Intentional} & Some intentional (sometimes subconscious) action/choice to incorrectly support an argument. & A woman decides to visit a certain doctor after only asking advice on the best doctors from ONE friend. \\ 
\midrule
\textit{False Causality} & Statement that jumps to a conclusion implying a causal relationship without supporting evidence. & Joan is scratched by a cat while visiting her friend. Two days later she comes down with a fever. Joan concludes that the cat's scratch must be the cause of her illness. \\ 
\midrule
\textit{False Dilemma} & Presenting only two options or sides when there are many options or sides. & Eat great food at our restaurant, or eat sad, boring meals at home. \\ 
\midrule
\textit{Faulty Generalization} & An informal fallacy wherein a conclusion is drawn about all or many instances of a phenomenon on the basis of one or a few instances of that phenomenon. & ALL teenagers are irresponsible. \\ 
\midrule
\textit{Fallacy of Credibility} & Attempts to disprove an argument by attacking the character of the speaker. & I hold a doctorate in theology, have written 12 books, and personally met the Pope.  Therefore, when I say that Jesus’ favorite snack was raisins dipped in wine, you should believe me. \\ 
\midrule
\textit{Fallacy of Logic} & An error in the logical structure of an argument. & If the state can require car seats for small children and infants, they can just as easily require mothers to breast-feed instead of using formula. \\
\midrule
\textit{Circular Reasoning} & When the end of an argument comes back to the beginning without having proven itself. & George Bush is a good communicator because he speaks effectively. \\ 
\midrule
\textit{Equivocation} & When a key term or phrase in an argument is used in an ambiguous way, with one meaning in one portion of the argument and then another meaning in another portion of the argument. & I don’t understand why you’re saying I broke a promise. I said I’d never speak again to my ex-girlfriend. And I didn't. I just sent her a quick text. \\ 
\bottomrule
\end{tabular}
\caption{Examples for fallacious arguments belonging to different classes covered in our work.
}
\label{tab:fallacy-examples}
\end{table*}

\begin{table*}[h]
\centering
\small
\begin{tabular}{l|rrrrrr|rr}
\toprule
Fallacy Type &  \textit{Counterarg.} &  \textit{Explanations} &  \textit{Goals} &  \textit{Structure} &   \textit{Text} & baseline & \# test & \# train \\
\midrule
\textit{Ad Hominem} &    0.781 &   0.732 &  0.759 &      0.756 &  \textbf{0.825} & 0.596 & 39 & 185 \\
\textit{Ad Populum}  &    \textbf{0.900} &         0.875 &  0.857 &      0.870 &  0.852 & 0.812 & 31 & 144\\
\textit{Appeal to Emotion}  &    \textbf{0.723} &         0.612 &  0.680 &      0.608 &  0.596 & 0.426 & 23 & 109\\
\textit{Circular Reasoning}  &    0.739 &         0.730 &  0.750 &      0.760 &  \textbf{0.800} & 0.524 & 23 & 110\\
\textit{Equivocation}  &    0.285 &         0.333 &  0.307 &      0.333 &  \textbf{0.352} & 0.000 & 7 & 32\\
\textit{Fallacy of Credibility}  &    \textbf{0.666} &         0.562 &  0.600 &      0.606 &  0.571 & 0.400 & 19 & 89\\
\textit{Fallacy of Extension}  &    0.705 &         0.702 &  0.666 &      0.666 &  \textbf{0.750} & 0.482 & 18 & 80\\
\textit{Fallacy of Logic}  &    \textbf{0.714} &         0.604 &  0.666 &      0.666 &  0.708 & 0.322 & 22 & 101\\
\textit{Fallacy of Relevance}  &    0.634 &         0.590 &  0.590 &      \textbf{0.651} &  0.553 & 0.512 & 22 & 102\\
\textit{False Causality}  &    \textbf{0.827} &         0.763 &  0.769 &      0.771 &  0.792 & 0.596 & 28 & 132\\
\textit{False Dilemma}  &    0.882 &         0.769 &  0.829 &      \textbf{0.894} &  0.857 & 0.800 & 19 & 87\\
\textit{Faulty Generalization}  &    0.699 &         0.598 &  0.656 &      0.666 &  \textbf{0.702} & 0.656 & 60 & 281 \\
\textit{Intentional}  &    0.628 &         0.545 &  0.578 &      \textbf{0.648} &  0.550 & 0.482 & 20 & 92\\
\bottomrule
\end{tabular}
\caption{
Per-class analysis of the model's performance using different case representations. The last two columns are the number of test data points, and training data points before augmentation.
}
\label{tab:results-per-class-all-case-representations}
\end{table*}

Due to the imbalance issue with different types of fallacies (see Table \ref{tab:results-per-class-all-case-representations}), we augment the dataset using common text augmentation techniques to have 281 arguments for each fallacy type. We tried back-translation, and substitution of entities in the arguments with their synonymous terms, but finally given more coherent samples created using substitution, we used substitution with synonymous terms using transformer embeddings. In this setting, we use transformer embeddings extracted from RoBERTa \cite{roberta} to replace words with their most similar candidates based on the extracted embeddings.

\section{Prompts}
\label{appendix:prompts}

We use prompting in two scenarios: to predict the fallacy type for an argument using Codex in a few-shot setting, and to extract the case representations using both ChatGPT and Codex combining zero-shot and few-shot prompting. 

\textbf{Extracting Fallacy Types.}
We use Codex in a few-shot setting to predict fallacy types for arguments and use these predictions as a baseline in our experiments. We do not run the same experiment using ChatGPT due to the limits imposed on its usage. The prompt we use to extract predictions from Codex contains all the thirteen possible classes of fallacies shown as a \textit{Python List} followed by one example with its correct prediction per class as shown below. \\

\begin{minipage}{\textwidth}
\begin{center}
\begin{addmargin}[2em]{2em}

\textit{classes = ['fallacy of logic', 'circular reasoning', }\\
\textit{'appeal to emotion', 'intentional',} \\
\textit{'faulty generalization', 'fallacy of extension',}\\
\textit{'false dilemma', 'ad populum', 'ad hominem', }\\
\textit{'false causality', 'equivocation', }\\
\textit{'fallacy of relevance', 'fallacy of credibility']}\\
--------\\
(
\begin{addmargin}[2em]{2em}
\textit{plain text of the argument}\\
\textit{correct fallacy type}\\
\#\#\#\\
\end{addmargin}
) $\times 13$ \\
\textit{plain text of the argument}\\\
\end{addmargin}
\end{center}
\end{minipage}

\textbf{Extracting Case Representations.}
We use prompting to get case representations in two steps: first, to get limited high-quality case representations, we use ChatGPT in a zero-shot manner; second, to extract case representations in high volume for all the arguments in our dataset, we prompt Codex in a few-shot setting including the sample responses extracted from ChatGPT.

To get the representations from ChatGPT, we use the prompts shown in Table \ref{tab:prompts-chatgpt}. We proofread the representations manually to verify their correct structure and soundness of the representation. Since we include these responses in the prompt we use in the second pass, we make sure they do not contain any statement about the logical fallacy that an argument contains. 
The described procedure is done for \textit{counterarguments}, \textit{explanations}, and \textit{goals}. However, for the \textit{structure} of the arguments, we manually write down the \textit{structure} of five sample arguments to be included in the prompt for the next step. We define the \textit{structure} of the argument as a statement in which the content words are replaced with symbols to serve as an abstraction from the plain text. In this representation, the same entities must have the same symbols, and also the named entities and content words that are not helping the model to recognize an argument as a logical fallacy are also replaced with symbols. For instance, the argument \textit{People either like coffee or hate it} will have a \textit{structure} as \textit{People either like x or hate x}.

Having five samples from different arguments with their \textit{counterarguments}, \textit{explanations}, \textit{structure}, and \textit{goals}, to get representations for the rest of the arguments we create a prompt filling it in with five samples developed in the previous stage and prompt Codex in the following way:\\

\begin{minipage}{\textwidth}
\begin{center}
\begin{addmargin}[2em]{2em}
(
\begin{addmargin}[2em]{2em}
\textit{prompt}:\\
\textit{response}\\
\#\#\#
\end{addmargin}
) $\times$ 5 \\
\textit{prompt with a new argument}:\\\
\end{addmargin}
\end{center}
\end{minipage}

\noindent Note that the \textit{prompt} is the same prompt we used for ChatGPT in Table \ref{tab:prompts-chatgpt} and the \textit{responses} are the high-quality responses we extracted from ChatGPT.

\begin{table*}[h]
\centering
\small
\begin{tabular}{p{0.15\textwidth} | p{0.33\textwidth} | p{0.47\textwidth}}
\toprule
\textbf{Representation} & \textbf{Prompt} & \textbf{Sample Answer} \\
\midrule
\textit{Explanations} & \underline{Analyze the argument} "guys are messy and unclean, but are really hard workers" & It's not appropriate to make generalizations about people based on their gender. The argument "guys are messy and unclean, but are really hard workers" suggests that all men are messy and unclean, which is not true. Additionally, the argument implies that being messy and unclean is at odds with being a hard worker, which is also not necessarily true. It's important to remember that every person is an individual with their own unique characteristics, and it's not fair to stereotype people or make assumptions about them based on their gender. \\
\midrule
\textit{Goals} & \underline{Express the goal of the argument} "The two courses I took at UWI were not very interesting. I don't think its a good university." & It's possible that the goal of the sentence "The two courses I took at UWI were not very interesting. I don't think its a good university." is to express the speaker's personal feelings and opinions about their academic experience at The University of the West Indies (UWI). The sentence could be communicating the speaker's dissatisfaction with UWI to someone else, or it may simply be a way for the speaker to process their own thoughts and feelings. \\
\midrule
\textit{Counterarg.} & \underline{Represent the counterargument to the argument} "four out of five dentists agree that brushing your teeth makes your life meaningful" & There is no evidence to support the claim that brushing your teeth makes your life meaningful. In fact, there are many other factors that contribute to a meaningful life, and it's not fair to suggest that brushing your teeth is a necessary or important factor. \\
\bottomrule
\end{tabular}
\caption{
Prompts applied on ChatGPT to extract different representations for each argument (The fixed part of the prompt is \underline{underlined}).
}
\label{tab:prompts-chatgpt}
\end{table*}

\section{Additional Analysis}
\label{sec:more-analysis}

\textbf{Effect of Different Representations.}
To develop a better intuition into the difference between the case representations, we perform a per-class analysis using different representations. 
The results of this per-class analysis are shown in Table \ref{tab:results-per-class-all-case-representations}. We observe different rankings of case representations within different fallacy types. In five out of thirteen classes, namely, \textit{Ad Hominem, Circular Reasoning, Equivocation, Fallacy of Extension}, and \textit{Faulty Generalization}, the plain \textit{text} of the arguments outperforms the other representations. In these classes, the additional information from external representations is not perceived as useful and is rather distracting to the model, although still helps CBR to outperform vanilla language models. In \textit{Ad Populum, Appeal to Emotion, Fallacy of Credibility, Fallacy of Logic,} and \textit{False Causality}, \textit{counterarguments} turn out to be more effective, which signifies the importance of considering \textit{counterarguments} when approaching these fallacies. On the other hand, \textit{structure} of the arguments are helping CBR the most in \textit{Fallacy of Relevance, False Dilemma, } and \textit{Intentional} fallacies. This observation aligns well with the fact that these fallacies have a distinguishable look if one looks at them from a more abstract perspective. Although in none of the fallacy types, the \textit{goals} of the arguments nor \textit{explanations} about them can outperform other representations, they are not always the last in the ranking and can outperform the \textit{text} or \textit{counterarguments} that are most helpful in general.

Alluding to the complexity of fallacy types and the fact that classifying logical fallacies even for great philosophers has always been a challenge culminating in different taxonomies of fallacies \cite{Aristotle1989-jz,barker1965elements,Ireneous_Nakpih_2020}, it is natural that it is impossible to approach all fallacies using only one point of view or representation. Instead, depending on the fallacy type and the type of logical error that we are dealing with, different perspectives are needed, which we provide using different representations of the arguments. 

\textbf{Effect of Attention Mechanism.}
We investigate the sensitivity of the best-performing CBR model based on ELECTRA to the attention mechanism located in the Adapter. Table \ref{tab:results-with-without-attention} demonstrates the performance of this model with and without using the attention mechanism. Note that in the case where we don't use the attention mechanism, the embedding of the current case alongside previous cases $E_S$ directly gets fed to the classifier. Our results show that the model that uses the attention mechanism consistently outperforms the model that comes without the attention mechanism. This trend can be seen consistently regardless of the case representation. These results show that the attention mechanism is an essential part of our framework, and the focus on the previous cases should be adjusted so that the best results can be obtained.

\begin{table*}[h]
\centering
\small
\begin{tabular}{@{}llrrrrrr@{}}
\toprule
& &
\multicolumn{3}{c}{ \textbf{LOGIC}} & \multicolumn{3}{c}{ \textbf{LOGIC Climate}} \\
\cmidrule(lr){3-5} \cmidrule(lr){6-8}
\textbf{Representation} & \textbf{Attention} & P & R & F1  & P & R & F1\\
\midrule
Text 
& \textit{w}         &     \textbf{0.655} &  \textbf{0.634} &  \textbf{0.635} &             \textbf{0.317} &          \textbf{0.242} &      \textbf{0.242} \\
& \textit{w/o}     & 0.566 & 	0.560 &	0.555 & 	0.258 & 	0.196 & 	0.193  \\
\midrule
Counterarg. 
& \textit{w}          &     \textbf{0.663} &  \textbf{0.664} &  \textbf{0.657} &             \textbf{0.355} &          \textbf{0.254} &     \textbf{0.270} \\
& \textit{w/o}     & 0.579 & 	0.574 & 	0.567 & 	0.283 & 	0.184 & 	0.187  \\
\midrule
Goals
& \textit{w}          &     \textbf{0.646} &  \textbf{0.622} &  \textbf{0.621} &  \textbf{ 0.376} &  \textbf{0.217} &     \textbf{ 0.222} \\
& \textit{w/o}     & 0.578 & 0.568	& 0.568	& 0.261	& 0.188 & 	0.191 \\
\midrule
Structure
& \textit{w}         &     \textbf{0.634} &  \textbf{0.625} &  \textbf{0.618} & \textbf{0.375} &  \textbf{0.254} &      \textbf{0.269} \\
& \textit{w/o}     & 0.615  &  0.616  &  0.612  &  0.293	&  0.214	&   0.222 \\
\midrule
Explanations
& \textit{w}   & \textbf{0.605} & \textbf{ 0.580} &  \textbf{0.578} & \textbf{ 0.314} &  \textbf{0.242} & \textbf{0.237} \\
& \textit{w/o}     & 0.520 & 	0.496 & 	0.491 & 	0.212 & 	0.140 & 	0.128 \\
\bottomrule
\end{tabular}
\caption{
Performance of the model with and without the attention mechanism located in the Adapter for each case representation. This comparison was done using the best combination of parameters according to our previous experiments. We use one similar case from the case database and also exploit only $0.1$ of the case database performing these experiments. Also, among the underlying LMs, we use ELECTRA for this experiment.
}
\label{tab:results-with-without-attention}
\end{table*} 

\end{document}